# AUTOMATIC HOME-BASED SCREENING OF OBSTRUCTIVE SLEEP APNEA USING SINGLE CHANNEL ELECTROCARDIOGRAM AND SPO2 SIGNALS

Hosna Ghandeharioun

Department of Electrical and Biomedical Engineering,
Khorasan Institute of Higher Education, Mashhad, Iran

## ABSTRACT

*Obstructive sleep apnea (OSA) is one of the most widespread respiratory diseases today. Complete or relative breathing cessations due to upper airway subsidence during sleep is OSA. It has confirmed potential influence on Covid-19 hospitalization and mortality, and is strongly associated with major comorbidities of severe Covid-19 infection. Un-diagnosed OSA may also lead to a variety of severe physical and mental side-effects. To score OSA severity, nocturnal sleep monitoring is performed under defined protocols and standards called polysomnography (PSG). This method is time-consuming, expensive, and requiring professional sleep technicians. Automatic home-based detection of OSA is welcome and in great demand. It is a fast and effective way for referring OSA suspects to sleep clinics for further monitoring. On-line OSA detection also can be a part of a closed-loop automatic control of the OSA therapeutic/assistive devices. In this paper, several solutions for online OSA detection are introduced and tested on 155 subjects of three different databases. The best combinational solution uses mutual information (MI) analysis for selecting out of ECG and SpO2-based features. Several methods of supervised and unsupervised machine learning are employed to detect apnoeic episodes. To achieve the best performance, the most successful classifiers in four different ternary combination methods are used. The proposed configurations exploit limited use of biological signals, have online working scheme, and exhibit uniform and acceptable performance (over 85%) in all the employed databases. The benefits have not been gathered all together in the previous published methods.*

## 1. INTRODUCTION

Obstructive sleep apnea (OSA) is the most prevalent sleep-related breathing disorder worldwide [1]. Occasional episodes of airway lowering during sleep characterizes OSA [4]. If OSA remains unrecognised, it follows sudden changes in sympathetic neural activity, sleep fragmentations and pressure/heart-beat fluctuations. These side-effects may cause severe cardiovascular complications [5], type 2 diabetes [6], and psychiatric symptoms [7]. OSA is also established as a comorbidity to Covid-19 [2]. People affected by OSA are at risk for severe adverse outcomes of Covid-19 infection and close monitoring should be performed on them [3]. Hence the immediate detection and treatment of OSA is crucial. To detect OSA, the simultaneous evaluation of related clinical features and the visible signs of abnormal breathing during sleep is needed. [8]. The gold standard for scoring abnormal breathing during sleep is nocturnal polysomnography (PSG). The PSG-driven apnea-hypopnea index (AHI) identifies the OSA severity [9, 10]. AHI is derived through visual inspection according to the guidelines of the American Association of Sleep Medicine (AASM). It imposes a heavy work-load on the public health section [11]. Therefore, many automatic methods for pre-clinic detection of OSA have been suggested in the literature [12-28, 32, 33, 35, 36]. Various biological signals and machine learning methods are used. In some studies, the features of the electroencephalogram (EEG) are extracted from the inconsistencies between the right and left hemispheres [12] or tracking non-linear dynamics of EEG due to changes in sleep depth [13, 30, 31]. Single-channel ECG or simultaneous use of ECG plus saturated oxygen level of the blood in peripheral veins (SpO2) is also suggested in several studies due to easy signal acquisition and promising results [13, 14, 16-19, 23].

Most recently, OSA is diagnosed by the ECG-derived features and the newly popular deep learning methods. In deep learning techniques, the features are generally extracted/reduced during the learning algorithm, and no separate step is needed [32]. This advantage reduces the computational burden. However, for the training phase, computers with high processing ability and good data storage capacity are inevitable [33]. These hardware requirements do not suit home-based and portable applications. Convolutional neural networks (CNN) are also used for OSA detection based on nasal pressure signals [34]. Several supervised machine learning methods are suggested for OSA detection with a single channel ECG signal in [35]. The achieved results are satisfactory, yet reported in a small database.

In this study, several configurations for online OSA diagnosis are suggested. Use of one or two biological signals, automatic and on-line detection, and acceptable performance over several databases are the advantages of our proposed method, not integrated together in previous works to the knowledge of the author.

## 2. MATERIAL AND METHOD

Detection of respiratory events based on biological signals, and with the help of artificial intelligence is generally divided into several steps in the supervised strategies [28]. First the signal records are labeled as apnoeic or normal by an expert clinician. In the second step, some features are extracted from the recorded signals based on mathematical transforms and relations. Thirdly, a classifier is trained with the features of apnoeic and normal signals and finally if classifier is trained properly, it can classify newly recorded signals as normal or apnoeic with an acceptable accuracy.

We conducted this study based on three databases. The first two databases are public: St. Vincent, University College Dublin (UCD) database [36], eight subjects of Apnea-ECG database [37] whose data include more signals than one ECG channel. The third database is exclusively at our disposal and referred as "the Sina database" hereon. This database includes clinical records of the sleep laboratory of Ibn-e-Sina Hospital, Mashhad, Iran, from July 2012to May 2014. The study was approved by the ethics committee overlooking the research proposal (permission no.92/620792, date 2014/03/07). The PSG (model: Alice LE, part no. 1002387, Philips Respironics) recordings were conducted in baseline montage with16 channels on the 158 referred patients. Out of all participants, 134 subjects were diagnosed with OSA, and 24 healthy according to the International Classification of Sleep Disorders II (ICSD-II) [8]. We considered sleep apneas as ≥10 s of airflow break offs and hypopnea as a ≥3% of oxygen desaturation/or arousal after a 50% decline in the baseline airflow. Figure 1 shows a view of the polysomnographic sensor connections and the subject state in one of the Sina database records.

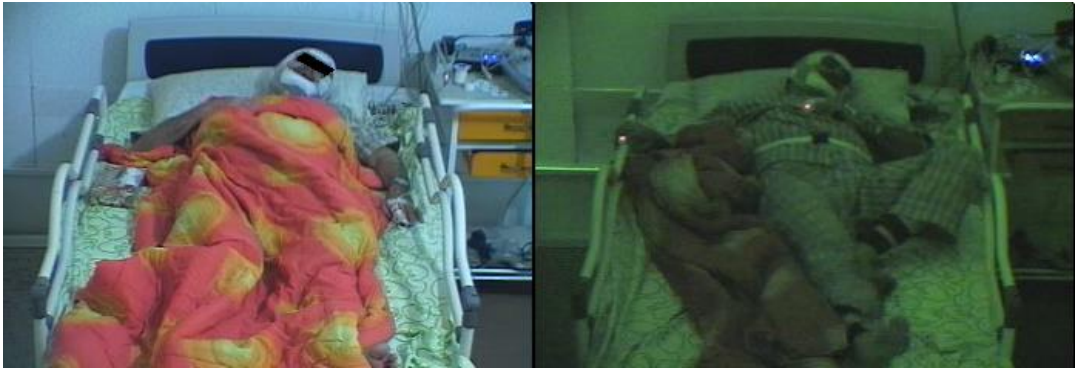

Figure 1. A view of the polysomnographic sensor connections and the subject in one of the Sina database records.

Electroencephalogram (EEG), Level of saturated oxygen in peripheral veins (SpO2), and air pressure/flow have a central role in the clinical definition of apnea. We refer to them as "the main signals". Other biological signals (e.g. Electrocardiogram, voice) are subsidiary and can act as "the auxiliary signals". Relying on the main signals for an OSA detection system is the first choice; however, the developed system must be more concise than PSG and perform a pre-clinic screening. Placing EEG electrodes on the scalp during sleep and pressure/flow sensors is rather obtrusive; besides, preparations and installation of electrodes and sensors are not straightforward for an ordinary user. For EEG acquisition and conditioning, a relatively expensive system is needed. The repeatability of the observed effects of OSA on EEG compared to SpO2 signal is also on debate [38, 39]. That is why generally EEG and air pressure/flow signals are excluded.

Among the auxiliary signals, ECG is gained more attention in the OSA detection methods. The effects of the apneas on ECG signal are well understood [4]. The ECG electrodes are installed simpler than EEG and less obtrusive than those of air pressure/flow signal. The apparent effect of respiratory events on ECG is called Cyclical Variation of Heart Rate (CVHR) [4]. The challenge of ECG-based detection systems is their lower specificity since their modulating factor is not a respiratory event only. The presence of cardiovascular problems can also have considerable effects on ECG. In the absence of OSA, these effects can increase the false positive detection rate. In practice, the number of false-negative detections also increases, and the sensitivity of the OSA detection method drops. A decrease in sensitivity is because the database usually includes subjects with OSA whose problem has been un-diagnosed for years, and lack of treatment has led to cardiovascular complexities for them [4]. Up to 90% of subjects affected by OSA are not aware of their problem and have not been treated yet [1].

More successful results are reported for SpO2-based detection methods compared to other single-channel detection systems. They have reasonable specificity and sensitivity, they can be performed in real-time, and they have non-obtrusive sensors; additionally, some of them are realized in smartphones and can serve as useful home-based systems [27, 40].

In this study, we consider PPG (and SpO2) from "the main signals", and ECG from "the auxiliary signals". Parallel use of these signals, covers their deficiencies and increases the overall accuracy, sensitivity, and specificity of the detection system [16]. The OSA detection based on ECG and SpO2 is more popular than other multi-channel detection systems due to simple sensor installation and powerful representation of respiratory events [13, 14, 16-19, 23].

## 2.1. Pre-processing and Noise Rejection

Considering the ECG sampling frequency is essential. The insufficient sampling frequency may negatively affect the resolution and the signal-to-noise ratio of the R-R time series [41, 42]. The UCD and the Apnea-ECG databases have less sampling frequency than the specified 250Hz value of the American National Standard Institute (ANSI), yet they are good benchmarks for the evaluation of automatic OSA detection methods. We have assumed that their subjects are carefully selected so that exceptions, where their sampling frequencies are insufficient for representing ECG behavior, are deleted [42]. The ECG signals of the exclusive database are also down-sampled to 250Hz.

To avoid the aliasing effects of non-integer fractional down-sampling, equating the UCD and the Apnea-ECG sampling frequencies is avoided [43]. For de-trending and noise rejection, the decimated lifting wavelet transform (DWT) algorithm [44] is employed [13]. The Daubechies (D4) wavelet is used with seven levels of decomposition. The R-R time series is extracted by the famous and robust method of Hamilton-Tompkins [45, 46]. Impulses more or less than 20% distant to the last normal R-R interval, those with more than 30% values in the R-S difference or with the negative R-S difference values are assumed to be a sign of ectopic or abnormal beat and omitted; the resulting signal is called the R-R tachogram [38].

Table 1. The SpO2 features in each 1-minute frame: $\{spo2_i\}_{i=1}^{60}$

| Name/ Definition |
| --- |
| The minimum value of the frame |
| The average value of the frame |
| The standard deviation of the frame |
| Sequential correlation coefficients [20] |
| Sequential mutual information [52] |
| Average value crossing points |
| The absolute value of the slope of the line fitted over SpO2 [20] |
| y-Intercept value of the line fitted over SpO2 [20] |
| Approximate entropy [53] |
| Sample entropy [53] |
| Lempel-Zive complexity measure [54] |
| Central tendency measure ( $CTM_r$ ) (r=0.25, 0.75, 0.5, 1)[54] |
| Delta measure ($\Delta$) [30] |
| Baseline [22] |
| odi2, odi3, odi4: The number of 2%,3%, and 4% desaturations to the baseline [30] |
| $ODIxy$: The number of desaturations more than or equal to $x$% lasting for $y$ seconds [30] |
| $ODISx$: The number of desaturations more than or equal to x% [22] |
| Time elapsed under saturation level x (%tsax ); x=70, 80,85, 90, 95) [30] |

## 2.2. Feature Extraction

We consider values below 50% and fluctuations more than 40% in two consecutive samples of SpO2 signal (in the sampling period of 1s) artifacts [16, 19]. We eliminate these values and their corresponding values of other PSG signals from the records (2 minutes of the Apnea-ECG database, 37 minutes of the UCD database, and 78 minutes of the exclusive database, totally equal to 1.9% of available data). The resulting signal is divided into non-overlapping 1-minute frames and is used for feature extraction. Table 1 summarizes the SpO2 features.

We process the ECG signal in 1-minute time windows. The R-R tachogram is extracted from ECG. It is not a result of uniform ECG sampling. The points of this time series are scattered non-uniformly across the time axis based on the time interval of consecutive beats. In frequency analysis of ECG signal, this crucial fact is usually ignored. The pre-assumption of the fast Fourier transform (FFT) is the uniform sampling of the signal under analysis; hence the FFT-based frequency analysis of the R-R tachogram and its dependents like the ECG-derived respiration (EDR) are not appropriate. Frequency analysis tools needless of the uniform sampling assumption like the Lomb-Scargle periodogram are good candidates for calculating quantities related to the heart rate variability (HRV) [50].

### 2.2.1. EDR Extraction

To find the best method for EDR extraction, we have done a quantitive evaluation among all EDR extraction methods based on single-channel ECG. Figure 2 summarizes all the suggested EDR extraction methods, based on their approach. Shaded blocks need at least two perpendicular ECG channels hence, immediately excluded from our choices [150-152].

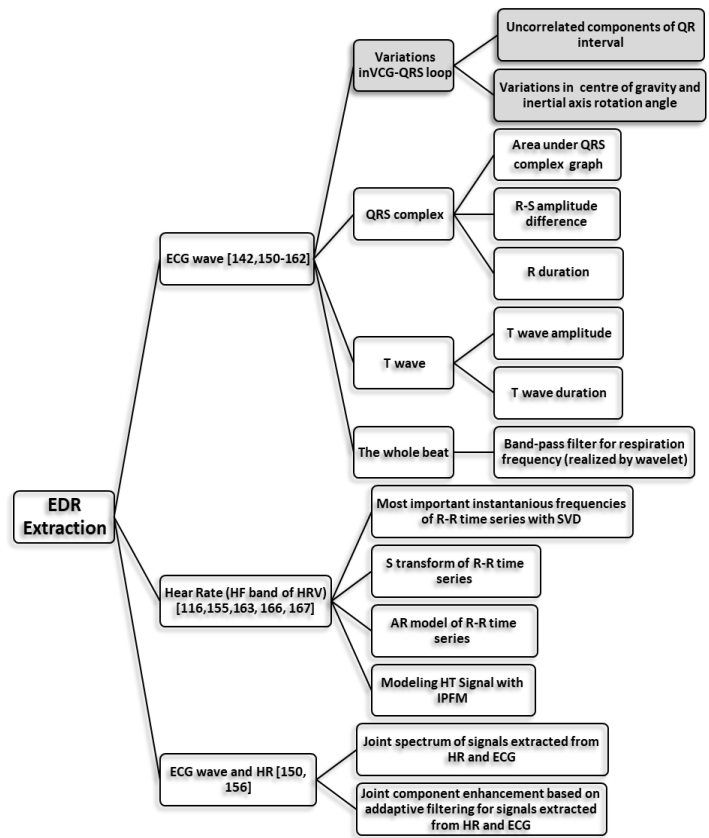

Figure 2. Categorization of EDR extraction methods; VCG (Vectorcardiogram), SVD (Singular value decomposition), AR(Auto-regressive) HR (Heart rate), HF (High frequency), HRV (Heart rate variability), IPFM (Integral pulse frequency modulation), The shaded blocks need at least two perpendicular ECG channels and are excluded from evaluation.

Due to excessive computational load, and need for off-line processing, approaches that integrate results of two separate methods (e.g. methods based on joint analysis of EDRs extracted separately from Hear rate (HR) and ECG signals[150,156]), methods than lie upon modelling (e.g. auto-regressive [155] or Integral pulse frequency modulation models [167]) or computationally intensive transforms (e.g. S-transform as a generalization of continuous wavelet transform [154] or singular value decomposition [153]) are also disregarded. We use approaches based on morphology of ECG waves [142,150-162]. Base of evaluation is comparing the dominant frequency of EDR signal and the dominant frequency of the respiratory signal (Oro-nasal air flow and thoracic and abdominal respiratory effort signals) simultaneously recorded with ECG. To reach the results, the FFT is performed on the three respiratory signals. The signals ate truncated into 30 s frames with 24 s overlap. The frequency components of the resulting periodogram are reported every 6 s and compared with dominant frequency of EDR signal.

EDR is extracted as a time series. To reach its dominant frequency, a 30 s window of the signal is selected and de-trended. A straight line is fitted over this 30-s window with the least-squares estimation and the difference is calculated. Then the EDR periodogram is computed by Lomb-Scargel method and dominant frequency is reported. Overlap of EDR windows is 24 s to have a new frequency every 6 seconds. The results are reported in table 2.

Table 2. The correlation coefficient between frequencies derived from respiratory signals and extracted EDRs. Numbers are reported as mean± Standard deviation.

| EDR Frequency | Respiration Frequency | | | | | | | | |
|---|---|---|---|---|---|---|---|---|---|
| | Oro-nasal air flow | | | Thoracic respiratory effort | | | Abdominal respiratory effort | | |
| | UCD | Apnea-ECG | Exclusive Database | UCD | Apnea-ECG | Exclusive Database | UCD | Apnea-ECG | Exclusive Database |
| Area under QRS complex graph | 0.85± 0.07 | 0.81 ± 0.07 | 0.90 ± 0.07 | 0.87 ± 0.08 | 0.85 ± 0.08 | 0.89 ± 0.08 | 0.90 ± 0.05 | 0.89 ± 0.05 | 0.92 ± 0.05 |
| R-S amplitude difference | 0.83 ± 0.09 | 0.80 ± 0.09 | 0.90 ± 0.09 | 0.83 ± 0.07 | 0.84 ± 0.07 | 0.88 ± 0.07 | 0.90± 0.05 | 0.88 ± 0.05 | 0.93 ± 0.05 |
| R wave duration | 0.82 ± 0.07 | 0.80 ± 0.07 | 0.88 ± 0.07 | 0.82 ± 0.08 | 0.80 ± 0.08 | 0.85 ± 0.08 | 0.84 ± 0.06 | 0.85 ± 0.06 | 0.90 ± 0.06 |
| T wave amplitude | 0.89 ± 0.07 | 0.9± 0.07 | 0.86 ± 0.07 | 0.88 ± 0.06 | 0.89 ± 0.06 | 0.85 ± 0.06 | 0.9± 0.05 | 0.9± 0.05 | 0.89 ± 0.05 |
| T wave duration | 0.91 ± 0.05 | 0.92 ± 0.05 | 0.89 ± 0.05 | 0.91 ± 0.07 | 0.91± 0.07 | 0.88 ± 0.04 | 0.95 ± 0.05 | 0.94 ± 0.07 | 0.89± 0.07 |
| Band-pass filtered R-R time series (0.2Hz-0.4 Hz, realized by wavelet) | 0.87 ± 0.08 | 0.85± 0.08 | 0.86 ± 0.08 | 0.85 ± 0.06 | 0.88 ± 0.06 | 0.87 ± 0.06 | 0.86 ± 0.05 | 0.88 ± 0.05 | 0.89 ± 0.05 |

We use the Lomb-Scargle periodogram and the DWT with Daubechies (D4) wavelet (with 18 levels of decomposition) to extract frequency-domain features of the R-R tachogram, and the EDR signals [44]. The ECG features are categorized as the time-domain, and the frequency-domain features in tables 2, 3 and 4.

Table 2. The time-domain ECG features

The R-R tachogram: $R\left(rr_{t_m}\right) = \{rr_i\}_{i=rr_{t_1}}^{rr_{t_m}}$, the EDR: $EDR(q) = \{edr_i\}_{i=1}^{q}$

| Definition | Name |
|---|---|
| $\overline{rr_t} = \dfrac{1}{m}\sum_{i=1}^{m} rr_{t_i}$ | Time window mid-time |
| M | length ECG |
| $\overline{rr} = \dfrac{1}{m}\sum_{i=1}^{m} rr_i$ | Average beat [115] |
| $NN50v1 = \sum_{i=2}^{m} U(|rr_i - rr_{i+1}| - 50ms)$  <br> U(.): step function | NN50-version 1 [115] |
| $NN50v2 = \sum_{i=1}^{m-1} U(|rr_{i+1} - rr_i| - 50ms)$  <br> U(.): step function | NN50-version 2 [115] |
| $pNN50v1 = \dfrac{NN50v1}{m}$ | pNN50-version 1 [115] |
| $pNN50v2 = \dfrac{NN50v2}{m}$ | pNN50-version 2 [115] |
| $S_{rr} = \sqrt{\dfrac{1}{m-1}\sum_{i=1}^{m}(rr_i - \overline{rr})^2}$ | Tachogram standard deviation |
| $SDSD = \sqrt{\dfrac{1}{m-1}\sum_{i=1}^{m}(rd_i - \overline{rd})^2}$  <br> $rd_i = rr_{i+1} - rr_i, \overline{rd} = \dfrac{1}{m-1}\sum_{i=1}^{m-1} rd_i$ | SDSD [115] |
| $RMSSD = \sqrt{\dfrac{1}{m-1}\sum_{i=1}^{m-1} rd_i^2}$ | RMSSD [115] |
| $r_k = \dfrac{\sum_{i=1}^{m}(rr_i - \overline{rr})(rr_{i+k} - \overline{rr})}{\sum_{i=1}^{m}(rr_i - \overline{rr})^2}$ | Sequential correlation coefficients [115] |
| $MI_k = \hat{I}(\{rr_i\}; \{rr_{i+k}\}) = \sum_{i=1}^{m} P_n(\{rr_i\}, \{rr_{i+k}\}) log\dfrac{P_n(\{rr_i\}, \{rr_{i+k}\})}{P_n(\{rr_i\})P_n(\{rr_{i+k}\})}$  <br> $P_n$: Probability distribution function | Sequential mutual information [319] |
| $AT_k = \dfrac{E\left((N_{i+1}(k) - N_i(k))^2\right)}{2E\left(N_{i+1}(k)\right)}$  <br> $N_i(k)$: Number of beats in the $i^{th}$ section of a k-second signal | Allan Factor [124] |
| $NEP_k = \dfrac{1}{\acute{m}-2}\sum_{i=2}^{\acute{m}-1}\left(1 - U((rr_i - rr_{i-1})(rr_{i+1} - rr_i))\right)$ | Number of Extreme Points [116] |
| $\overline{edr} = \dfrac{1}{q}\sum_{i=1}^{q} edr_i$ | Average EDR |
| $S_{edr} = \sqrt{\dfrac{1}{q-1}\sum_{i=1}^{q}(edr_i - \overline{edr})^2}$ | Standard Deviation EDR |

Table 3. The frequency-domain features of the R-R tachogram: $R\left(rr_{t_m}\right) = \{rr_i\}_{i=rr_{t_1}}^{rr_{t_m}}$

| Definition | Name |
|---|---|
| $S^2_{D_{rr}{}^k} = \sum_{i=1}^{l_{rr,k}} \left(d_{rr,i}{}^k - \overline{d_{rr}{}^k}\right)^2$ <br> $\overline{d_{rr}{}^k} = \frac{1}{l_{rr,k}} \sum_{i=1}^{l_{rr,k}} d_{rr,i}{}^k$ | Sample deviation of $\{D_{rr}{}^k\}_{k=2}^{17}$ |
| $S^2_{D_{rr}{}^{LF}} = \sum_{i=1}^{l_{rr,LF}} \left(d_{rr,i}{}^{LF} - \overline{d_{rr}{}^{LF}}\right)^2$ | Sample deviation of $\{D_{rr}{}^k\}_{k=2}^{17}$ (LF band) |
| $S^2_{D_{rr}{}^{HF}} = \sum_{i=1}^{l_{rr,HF}} \left(d_{rr,i}{}^{HF} - \overline{d_{rr}{}^{HF}}\right)^2$ | Sample deviation of $\{D_{rr}{}^k\}_{k=2}^{17}$ (HF band) |
| $P_{rr}{}^{VLF} = \int_{2\pi \times 0.04}^{2\pi \times 0.15} P_{rr}(\omega) d\omega$ <br> $P_{rr}(\omega)$ : Lomb-Scargel periodogram [348,86] | HRV Power spectrum (LF band) |
| $P_{rr}{}^{HF} = \int_{2\pi \times 0.15}^{2\pi \times 0.4} P_{rr}(\omega) d\omega$ | HRV Power spectrum (HF band) |
| $LF/HF = P_{rr}{}^{LF} / P_{rr}{}^{HF}$ | LF-HF power ratio in the HRV spectrum |
| $P_{rr}(\omega)\big|_{2\pi \times 0.04}^{2\pi \times 0.4}$ | Lomb-Scargel periodogram samples in LF-HF band |
| $\omega_{resp} = argmax(P_{rr}(\omega)\big|_{2\pi \times 0.15}^{2\pi \times 0.4})$ | Estimated respiration frequency (Dominant HF-band frequency of HRV) [86] |
| $respMag = max(P_{rr}(\omega)\big|_{2\pi \times 0.15}^{2\pi \times 0.4}) = P_{rr}(\omega_{resp})$ | Power at the dominant HF-band frequency of HRV |
| $respProb = Prob\left(P_{rr}(\omega_{resp})\right)$ | Probability of estimated respiration frequency occurrence with power $P_{rr}(\omega_{resp})$ |
| $\omega_{ProbMax} = argmax\left(Prob(P_{rr}(\omega)\big|_{2\pi \times 0.04}^{2\pi \times 0.4})\right)$ | Most probable frequency of the HRV spectrum |
| $ProbMax = max(Prob(P_{rr}(\omega)\big|_{2\pi \times 0.04}^{2\pi \times 0.4})) = Prob(P_{rr}(\omega_{ProbMax}))$ | Probability of $\omega_{ProbMax}$ occurrence with power $P_{rr}(\omega_{ProbMax})$ |
| $ProbMaxMag = P_{rr}(\omega_{ProbMax})$ | Power of the HRV spectrum at $\omega_{ProbMax}$ |

Table 4. The frequency-domain features of the EDR: $EDR(q) = \{edr_i\}_{i=1}^{q}$

| Definition | Name |
|---|---|
| $S^2_{D_{edr}{}^k} = \sum_{i=1}^{I_{edr,k}} \left(d_{edr,i}{}^k - \overline{d_{edr}{}^k}\right)^2$ $\overline{d_{edr}{}^k} = \frac{1}{I_{edr,k}} \sum_{i=1}^{I_{edr,k}} d_{edr,i}{}^k$ | Sample deviation of $\{D_{edr}{}^k\}_{k=2}^{17}$ |
| $S^2_{D_{edr}{}^{LF}} = \sum_{i=1}^{I_{edr,LF}} \left(d_{edr,i}{}^{LF} - \overline{d_{edr}{}^{LF}}\right)^2$ | Sample deviation of $\{D_{edr}{}^k\}_{k=5}^{17}$ (LF band) |
| $S^2_{D_{edr}{}^{HF}} = \sum_{i=1}^{I_{edr,VLF}} \left(d_{edr,i}{}^{HF} - \overline{d_{edr}{}^{HF}}\right)^2$ | Sample deviation of $\{D_{rr}{}^k\}_{k=2}^{4}$ (HF band) |
| $P_{edr}{}^{VLF} = \int_{2\pi \times 0.04}^{2\pi \times 0.15} P_{edr}(\omega) d\omega$ | EDR Power spectrum (LF band) |
| $P_{edr}{}^{HF} = \int_{2\pi \times 0.15}^{2\pi \times 0.4} P_{edr}(\omega) d\omega$ | EDR Power spectrum (HF band) |
| $LF/HF_{edr} = P_{edr}{}^{LF}/P_{edr}{}^{HF}$ | LF-HF power ratio in the EDR spectrum |
| $P_{edr}(\omega)\vert_{2\pi \times 0.04}^{2\pi \times 0.4}$ | Lomb-Scargel periodogram samples in LF-HF band |
| $\omega_{edr-resp} = argmax(P_{edr}(\omega)\vert_{2\pi \times 0.15}^{2\pi \times 0.4})$ | Dominant HF-band frequency of the EDR |
| $respMag_{edr} = max\left(P_{edr}(\omega)\vert_{2\pi \times 0.15}^{2\pi \times 0.4}\right) = P_{edr}(\omega_{edr-resp})$ | Power at the dominant HF-band frequency of the EDR |
| $respProb_{edr} = Prob\left(P_{edr}(\omega_{edr-resp})\right)$ | Probability of $\omega_{edr-resp}$ occurrence with power $P_{edr}(\omega_{edr-resp})$ |
| $\omega_{edr-ProbMax} = argmax\left(Prob(P_{edr}(\omega)\vert_{2\pi \times 0.04}^{2\pi \times 0.4})\right)$ | Most probable frequency of the EDR spectrum |
| $ProbMax_{edr} = max\left(Prob(P_{edr}(\omega)\vert_{2\pi \times 0.04}^{2\pi \times 0.4})\right) = Prob\left(P_{edr}(\omega_{edr-ProbMax})\right)$ | Probability of $\omega_{edr\text{-}ProbMax}$ occurrence with power $P_{edr}(\omega_{edr-ProbMax})$ |
| $ProbMaxMag_{edr} = P_{edr}(\omega_{edr-ProbMax})$ | Power of the HRV spectrum at $\omega_{edr\text{-}ProbMax}$ |

## 2.3. Feature Reduction

Most automatic OSA detection methods [11-13, 16-19, 27, 29] use no feature reduction or employ linear dependency and correlation-based strategies or principal component analysis (PCA) for feature selection. Dependency and mutual information (MI) proved to outperform linear methods of feature selection, especially in respiratory event detection [14, 31]. Feature selection can be performed by individual analysis of each feature [13, 21, 26]. It is also possible to define a measure to evaluate a subset of features [14, 16]. The first method speculates the inter-relations among features but, the second method searches for features with both the tightest relations with the class label and the loosest interaction with each other. We use the second strategy for feature reduction.

To calculate the mutual interactions, we consider MI rather than a simple statistical correlation. We select the features which have the highest MI with the class label (normal of apnoeic) and the least MI with each other. The approach to search the feature space is forward feature selection. In this approach, the subset of selected features is gradually built by adding single features to an initial null set [14, 54].

## 2.4. Classification

We employ nine classifiers in this study; support vector machines (SVM) [55], K nearest neighbors (KNN) [60], decision table [56], C4.5 [57] decision tree, reduced-error pruning tree (REPT) [58], functional trees [59], the meta-algorithm of adaptive boosting accompanied with the simple classifier of decision stump [60], and the meta-algorithm of bagging along with the alternating decision tree (ADT) [61]. The meta-algorithms make a new data set out of the primary data set and devise a new classifier for each set in one trial. These trials are repeated T times, and eventually, the results of the T classifiers are combined to achieve a more accurate result.

In this study, four classifier combination methods are also performed on a group of three binary classifiers. Combination methods are max probability, average probability, the product of probability, and majority voting [16].

## 3. RESULTS

Table 2 demonstrates the Pierson`s correlation coefficients between frequencies derived from respiratory signals and the extracted EDRs in each database. (8 subjects from Apnea-ECG database, 25 subjects of the UCD database, and 158 subjects of the exclusive database are considered). Separate evaluation in databases is due to their different sampling frequencies of ECG signals. The best performances are shaded. According to this table the EDR is extracted by the T wave duration method [51, 52] in the UCD and ECG-Apnea databases. We calculate the EDR with the help of the area under the QRS graph [53] in our exclusive database.

Higher correlation coefficients for the "area under the QRS graph" method comparing to approaches based on R wave amplitude and duration approach exhibits its priority and better performance. The area under graph is more stable and robust to noise [thesis161].

Relatively low sampling frequency of ECG signal in the UCD and Apnea-ECG database (less than 250Hz) makes the representation of the area under the graph inaccurate. The most rapid changes in ECG signal is occurred during systole and ejection of blood out of ventricles. This is the reason why these methods cannot represent respiratory frequency well especially during respiratory events. The best alternative is T wave EDR extraction. T wave is the result of slow repolarization of ventricles after systolic activity and is not affected by low sampling frequency of ECG signal [145 thesis]. Only when the T wave is not observed in one beat or inverted due to a cardiac problem, QRS area methods are considered [116thesis].

Table 5 demonstrates the selected features employing the MI measure. According to table 5, as the number of database subjects increases, the number of selected features also increases. There are several similarities between the selected measures; fewer ECG features are among the selected ones, mostly the time domain ECG features. This result is consistent with the previously published reports. Most of the selected features are based on the SpO2 signal, which indicates their power for the OSA detection. However, simultaneous use of the ECG and the SPO2 features enhances the performance of the OSA detection method [16].

Table 5. The selected features through forward feature selection based on the MI criterion. Name and definition of features stated in tables 1 to 4

| Number | Selected features | Database |
|---|---|---|
| 20 | MI$_3$, spo2$_{min}$, NEP$_1$, S$_{spo2}$, $MI_{spo2,1}$, $\Delta$, LZ$_{down}$, odi4, CTM$_{0.5}$, ODI55, tsa80, tsa85, tsa90, $S^2_{D_{rr}}$4, $P_{rr}{}^{HF}$, $S^2_{D_{edr}}$6, $P_{edr}{}^{LF}$, ‹samples 13$^{th}$ and 55$^{th}$ of $P_{rr}(\omega)$ sample 4$^{th}$ of $P_{edr}(\omega)$ | UCD |
| 18 | S$_{spo2}$, $MI_{spo2,1}$, $\Delta$, LZC$_{up}$, CTM$_{0.25}$, CTM$_{0.5}$, ODI55, tsa80, tsa85, tsa90, $S^2_{D_{rr}}$4, $S^2_{D_{edr}}$6, $P_{edr}{}^{LF}$, samples 11$^{th}$, 18$^{th}$, 22$^{th}$ and 55$^{th}$ of $P_{rr}(\omega)$ and sample 4$^{th}$ of $P_{edr}(\omega)$ | Apnea-ECG |
| 29 | Spo2$_{min}$, $\overline{spo2}$, S$_{spo2}$, $r_{spo2,2}$, ZC, ApEn, SpEn, LZC$_{up}$, $MI_{spo2,3}$, $MI_{spo2,4}$, $\Delta$, ODIS4, ODI23, ODI25, ODI31, ODI35, ODI51, ODI53, ODI55, odi3, odi4, odi5, tsa95, tsa85, tsa80, CTM$_{0.5}$, CTM$_{0.75}$, CTM$_1$, ‹ sample 4$^{th}$ of $P_{edr}(\omega)$ | Exclusive database |

Table 6 illustrates the performance of our real-time detection method in each of the databases. We obtain the results from a system equipped with Windows 10 Pro, version 1511, the Intel processor Core i7CPU M640@2.8GHz and a RAM of 4GB. All the classifiers are realized in Java language. Evaluation is 10-fold cross-validation.

In some references, only the classifier`s training time is reported [16]. This parameter is not enough to represent the total computational burden of the suggested method. In some previous works, the processing time is reported for a specified number of samples [14]. In our study, "the processing time for a fixed number of data samples" is not an accurate measure since several databases with different ECG sampling rates are observed.

Table 6. The performance of the suggested detection method in each of the databases: DT (Decision Table), REPT (Reduced-Error Pruning Tree), FT (Functional tree), AB+DS (Adaptive boosting + decision stump), B+ REPT (Bagging + REPT), B+ADT (bagging + alternating decision tree), AECG (Apnea-ECG database), EX (Exclusive database). Maximums in each column are shaded.

| Processing time for 10 frames | | | Accuracy (%) | | | Specificity (%) | | | Sensitivity (%) | | | Class ifier |
|---|---|---|---|---|---|---|---|---|---|---|---|---|
| EX | AECG | UCD | EX | AECG | UCD | EX | AECG | UCD | EX | AECG | UCD | |
| 19 | 11.8 | 11.9 | 88.3 | 95.3 | 82 | 91 | 89.8 | 93 | 80.9 | 96.68 | 81.02 | SVM |
| 7 | 2.98 | 2.09 | 82.9 | 90.4 | 82 | 84.7 | 94 | 83 | 80.01 | 89 | 80.5 | KNN |
| 5.68 | 3.001 | 2.503 | 82 | 83.7 | 82 | 83 | 84.9 | 82 | 82.9 | 83 | 82.9 | DT |
| 4.001 | 1.45 | 1.076 | 82 | 85.6 | 81.7 | 86.1 | 89 | 85 | 73 | 82.1 | 72 | C4.5 |
| 2.32 | 1.045 | 1.002 | 84.6 | 91.6 | 83.6 | 84.9 | 92.6 | 84 | 82.9 | 83.5 | 81.5 | REP T |
| 9.867 | 4.7 | 4.345 | 80 | 88.8 | 79.8 | 82 | 90.7 | 81.7 | 73 | 81.4 | 71.5 | FT |
| 2.383 | 1.32 | 1.205 | 92.6 | 87.3 | 79 | 93.3 | 79.3 | 78 | 89.9 | 92.6 | 88 | AB+ DS |
| 4.794 | 2.97 | 2.164 | 88.5 | 91 | 85 | 89.9 | 92.2 | 86.3 | 82.1 | 89 | 81.03 | B+ REP T |
| 29.9 | 15 | 13.98 | 85.6 | 95 | 84.5 | 85 | 95 | 83 | 86.78 | 89.9 | 85 | B+A DT |
| 18 | 9 | 8.99 | 55.1 | 63 | 57 | 55.8 | 57 | 54.3 | 55.01 | 65 | 59 | SOM |
| 10 | 5.7 | 4.897 | 35.6 | 38.6 | 34.5 | 33 | 37 | 33 | 38.4 | 40.1 | 37.3 | K-mean s |

Observing the processing time in table 6 reveals that the parameter value does not exceed 1s in the UCD and Apnea-ECG databases and 2s in our exclusive database. These margins are the minimum time needed for pre-processing and feature extraction at the specified sampling frequencies. Smaller values for processing times belong to the Apnea-ECG database with the lowest number of data points. The processing time for our exclusive database is the highest of all, nearly two times the minimum value. Regarding this quantity, two classifiers have the highest computational burden; the ADT and the SVM. The processing time for the SVM is more than two times higher than the others`. For the real-time OSA detection, these computationally intensive classifiers are not chosen despite their high classification ability.

Accuracy, sensitivity, and specificity in all the databases are satisfactory but, slightly better in the Apnea-ECG database compared to the others. The two unsupervised classifies (the SOM and the K-means) do not exhibit acceptable results. Best sensitivity, but the worst specificity/accuracy belongs to adaptive boosting accompanied with the decision stump. On the other hand, bagging along with REPT achieves the best accuracy and specificity at the price of degrading sensitivity.

To reach a method with acceptable sensitivity and specificity, the combination routines declared in section 2.4 are used to fuse a group of three classifiers. Because the "boosting with the decision stump" and the "bagging along with the REPT" have better performances than others, they are the two fixed members of the group. The third member is chosen from the rest of the classifiers. We exclude the SVM and the "bagging with ADT" due to excessive computational load, so five options remain. These classifiers shape five different classifier groups to be fused. The classifier combination results are reported in tables 7 to 9.

According to tables 7 to 9, performance is nearly equal in all databases (slightly better performance for the Apnea-ECG database). Combining the classifiers, balances the performance measures in values around 80%. The most successful combination happened when the third group member is the KNN or the decision tree. In these cases, all the measures of performance, including sensitivity, specificity, and accuracy, have achieved values of more than 85%. These results outperform all the suggested methods to date [13, 14, 16, 19, 32, 35]. The principal difference between the KNN and the decision tree lies in their nature. KNN benefits from slow, moment-based training. It is appropriate for subject-dependant applications, in which models are built and tested with the same data. In subject-dependant applications each classifier model should be trained (i.e. updated) with the user data before utilization. On the other hand, the decision table is suitable for subject-independent applications where the classifier model is trained with a database of several subjects before being tested by the user.

Surveying the processing time shows that this quantity is approximately equal to the sum of the processing time needed for each classifier of the group. There is no distinguished difference between different combination routines. It is worth saying that combination methods based on probability need the sensitivity and the specificity of the classifier to weigh their decisions. This issue entails a more complex online realization than that of majority voting. Therefore, in online realization, the majority voting method will suffice.

Table 7. The performance of the suggested classifier combination detection method in the UCD database. Three classifiers are combined with four different methods (MP: Maximum probability, PP: Probability product, AP: Average probability, MV: Majority voting). Other abbreviations are similar to table 6. The two highest values in each column are shaded.

| Processing time for 10 frames | | | | Accuracy (%) | | | | Specificity (%) | | | | Sensitivity (%) | | | | 3rd classifier |
|---|---|---|---|---|---|---|---|---|---|---|---|---|---|---|---|---|
| MV | AP | PP | MP | MV | AP | PP | MP | MV | AP | PP | MP | MV | AP | PP | MP | |
| 4.36 | 4.68 | 4.47 | 4.4 | 85.28 | 86.12 | 86.2 | 86.12 | 85.25 | 86.03 | 86.16 | 86.07 | 87.55 | 87.41 | 87.19 | 85.87 | KNN |
| 4.55 | 4.65 | 4.76 | 4.869 | 85 | 85.68 | 85.7 | 85.64 | 84.16 | 85.35 | 85.42 | 85.47 | 87.61 | 86.68 | 86.57 | 86.14 | DT |
| 4 | 3.92 | 3.79 | 3.963 | 81.81 | 82.12 | 82.17 | 82.02 | 81.25 | 82.03 | 82.16 | 82.07 | 83.55 | 82.41 | 82.19 | 81.87 | C4.5 |
| 3.65 | 3.56 | 3.39 | 3.245 | 81 | 81.68 | 81.70 | 81.64 | 80.16 | 81.35 | 81.42 | 81.47 | 83.61 | 82.68 | 82.57 | 82.14 | REPT |
| 5.21 | 5.34 | 5.63 | 5.56 | 81.03 | 80.95 | 80.98 | 80.96 | 80.43 | 80.48 | 80.57 | 80.69 | 82.9 | 82.41 | 82.25 | 81.82 | FT |

Table 8. The performance of the suggested classifier combination detection method in the Apnea-ECG database. Three classifiers are combined with four different methods (MP: Maximum probability, PP: Probability product, AP: Average probability, MV: Majority voting). Other abbreviations are similar to table 6. The two highest values in each column are shaded.

| Processing time for 10 frames | | | | Accuracy (%) | | | | Specificity (%) | | | | Sensitivity (%) | | | | 3rd classifier |
|---|---|---|---|---|---|---|---|---|---|---|---|---|---|---|---|---|
| MV | AP | PP | MP | MV | AP | PP | MP | MV | AP | PP | MP | MV | AP | PP | MP | |
| 5.68 | 5.634 | 5.555 | 5.29 | 85.38 | 86.2 | 86.27 | 86.15 | 85.34 | 86.1 | 86.23 | 86.17 | 87.6 | 87.5 | 87.2 | 86 | KNN |
| 5.23 | 5.125 | 5.34 | 5.291 | 85.02 | 85.7 | 85.8 | 85.2 | 84.2 | 85.39 | 85.48 | 85.5 | 86.70 | 86.73 | 86.6 | 86.23 | DT |
| 3.99 | 3.7 | 3.65 | 3.49 | 82.4 | 82.25 | 82.2 | 82.23 | 82.33 | 82.1 | 82.2 | 82.1 | 83.6 | 82.5 | 82.2 | 81.94 | C4.5 |
| 3.025 | 3.068 | 3.128 | 3.11 | 81 | 81.71 | 81.75 | 81.7 | 80.2 | 81.38 | 81.49 | 81.5 | 83.69 | 82.7 | 82.6 | 82.15 | REPT |
| 5.969 | 5.79 | 5.87 | 5.9 | 81.1 | 81.2 | 81.01 | 81 | 80.57 | 80.6 | 80.7 | 81 | 83 | 82.5 | 82.31 | 81.91 | FT |

Table 9. The performance of the suggested classifier combination detection method in the exclusive database. Three classifiers are combined with four different methods (MP: Maximum probability, PP: Probability product, AP: Average probability, MV: Majority voting). Other abbreviations are similar to table 6. The two highest values in each column are shaded.

| Processing time for 10 frames | | | | Accuracy (%) | | | | Specificity (%) | | | | Sensitivity (%) | | | | 3rd classifier |
|---|---|---|---|---|---|---|---|---|---|---|---|---|---|---|---|---|
| MV | AP | PP | MP | MV | AP | PP | MP | MV | AP | PP | MP | MV | AP | PP | MP | |
| 12.05 | 11.81 | 11.6 | 12 | 85.32 | 86.03 | 86.13 | 86 | 85.24 | 86 | 86.11 | 86 | 87.23 | 87.35 | 87.1 | 85.67 | KNN |
| 10.43 | 10.54 | 10.68 | 10.56 | 84.9 | 85.7 | 85.7 | 85.48 | 84.14 | 85.30 | 85.34 | 85.5 | 87.5 | 86.65 | 86.6 | 86 | DT |
| 9.34 | 9.47 | 9.24 | 9.04 | 81.78 | 82.10 | 82.13 | 81.95 | 81.20 | 82 | 82.14 | 82.01 | 83.51 | 82.13 | 82.17 | 81.85 | C4.5 |
| 7.349 | 7.367 | 7.489 | 7.32 | 82.1 | 81.68 | 81.67 | 81.64 | 81.55 | 81.33 | 81.43 | 81.46 | 83.6 | 82.7 | 82.55 | 82.1 | REPT |
| 15.004 | 14.96 | 14.62 | 14.86 | 81.01 | 80.87 | 80.93 | 80.92 | 80.32 | 80.36 | 80.44 | 80.59 | 82.8 | 82.3 | 82.15 | 81.72 | FT |

## 4. CONCLUSIONS

In this study, several configurations for online detection of the OSA are suggested. The advantages of the proposed method are: exploiting only two channels of biological signals, automatic and real-time detection, and uniform acceptable performance over several databases (over 85%). To date, no other study has achieved all these merits together. Acceptable performance in well-known databases is due to classifiers that do not possess database-related parameters (e.g. sampling frequency of signals). The classifiers have covered deficiencies of each other in a combinational configuration. To reach the best result, the most successful classifiers are combined in groups of three members with four different combination methods. The features are also calculated and selected considering generality; in frequency-domain analysis, the refined Lomb-Scargle periodogram is used to care for the inherent non-uniform sampling of the R-R tachograms and unequal sampling frequency of the ECG signal in different databases [50]. Feature selection is based on the MI. The MI measure considers non-linear correlations among features and selects effective features to decrease the computational burden of the classifiers and avoid over-fitting problems.

On the other hand, the MI feature reduction has an important impact on the family of decision tree classifiers. MI-based feature selection accompanied by decision tree classifiers, avoids the classifier sensitivity to MI-biased estimates. In other words, the decision-tree classifiers may be misled by a fake replica of a feature with more marginal samples and higher maximum entropy value [62]. Selection of the more appropriate feature with an entropy-normalised MI estimator is helpful [62, 63].

ACKNOWLEDGEMENTS

The author appreciates the cooperation of the Ibn-e-Sina Hospital sleep laboratory at Mashhad University of Medical Sciences.

## AUTHOR

**Dr. Hosna Ghandeharioun** received her B.S. degree in electrical engineering from Ferdowsi University of Mashhad (FUM), Iran, in 2003 and an M.S. degree in Biomedical Engineering from Iran University of Science & Technology, Iran in 2006. She received her Ph. D. degree in electrical engineering from FUM in 2016. Now She works as an assistant professor at the electrical and biomedical Engineering dept. of Khorasan Institute of Higher Educations. Her research interests are biological signal processing and mobile health.

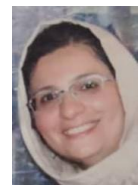